\documentclass[journal]{IEEEtran}

\usepackage[T1]{fontenc}
\usepackage[latin9]{inputenc}
\usepackage{color}
\usepackage{array}
\usepackage{float}
\usepackage{multirow}
\usepackage{amsmath}
\usepackage{amssymb}
\usepackage{stackrel}
\usepackage{graphicx}

\makeatletter

\providecommand{\tabularnewline}{\\}
\floatstyle{ruled}
\newfloat{algorithm}{tbp}{loa}
\providecommand{\algorithmname}{Algorithm}
\floatname{algorithm}{\protect\algorithmname}

\newenvironment{lyxcode}
	{\par\begin{list}{}{
		\setlength{\rightmargin}{\leftmargin}
		\setlength{\listparindent}{0pt}
		\raggedright
		\setlength{\itemsep}{0pt}
		\setlength{\parsep}{0pt}
		\normalfont\ttfamily}%
	 \item[]}
	{\end{list}}

\usepackage{algorithm}
\usepackage{algpseudocode}
\usepackage{flushend}

\usepackage[numbers,sort&compress]{natbib} 

\usepackage{breakurl}

\usepackage{colortbl}
\definecolor{grey}{cmyk}{0, 0, 0, 0.4}
\definecolor{lightblue}{cmyk}{0.25, 0.06, 0, 0.1}
\definecolor{palegreen}{cmyk}{0.39, 0, 0.39, 0.02}
\definecolor{lightgrey}{cmyk}{0, 0, 0, 0.17}

\renewcommand{\ALG@name}{Routine}

\@ifundefined{showcaptionsetup}{}{%
 \PassOptionsToPackage{caption=false}{subfig}}
\usepackage{subfig}
\makeatother

\usepackage{babel}
\begin{document}
\title{An Online Spatial-Temporal Graph Trajectory Planner for Autonomous
Vehicles}
\author{Jilan Samiuddin, Benoit Boulet and Di Wu}
\IEEEaftertitletext{\noindent \thanks{Manuscript received $\qquad\qquad\qquad$; revised $\qquad\qquad\qquad$;
accepted $\qquad\qquad\qquad$; This work was supported by Quebec\textquoteright s
Fonds de recherche Nature et technologies. (Corresponding authors:
Jilan Samiuddin, Di Wu.)\\The authors are with the Intelligent Automation
Lab of the Centre for Intelligent Machines and the Department of Electrical
and Computer Engineering, McGill University, Montreal, QC H3A 0E9,
Canada (email: jilan.samiuddin@mail.mcgill.ca, benoit.boulet@mcgill.ca,
di.wu5@mcgill.ca).}}
\maketitle
\begin{abstract}
The autonomous driving industry is expected to grow by over 20 times
in the coming decade and, thus, motivate researchers to delve into
it. The primary focus of their research is to ensure safety, comfort,
and efficiency. An autonomous vehicle has several modules responsible
for one or more of the aforementioned items. Among these modules,
the trajectory planner plays a pivotal role in the safety of the vehicle
and the comfort of its passengers. The module is also responsible
for respecting kinematic constraints and any applicable road constraints.
In this paper, a novel online spatial-temporal graph trajectory planner
is introduced to generate safe and comfortable trajectories. First,
a spatial-temporal graph is constructed using the autonomous vehicle,
its surrounding vehicles, and virtual nodes along the road with respect
to the vehicle itself. Next, the graph is forwarded into a sequential
network to obtain the desired states. To support the planner, a simple
behavioral layer is also presented that \textcolor{black}{determines
kinematic constraints} for the planner. Furthermore, a novel potential
function is also proposed to train the network. Finally, the proposed
planner is tested on three different complex driving tasks, and the
performance is compared with two frequently used methods. The results
show that the proposed planner generates safe and feasible trajectories
while achieving similar or longer distances in the forward direction
and comparable comfort ride.
\end{abstract}

\begin{IEEEkeywords}
Autonomous driving, graph neural network, online trajectory planner,
spatial, temporal
\end{IEEEkeywords}

\section{Introduction}

Autonomous vehicles have the potential to improve overall transportation
mobility in terms of safety and efficiency. The module which is primarily
responsible for planning safely the motion of the vehicle through
traffic is the trajectory planner. The trajectory planner is a vast
and long-researched area using a wide variety of methods such as different
optimization techniques, artificial intelligence and machine learning
\cite{claussmann2019review}. In this work, we propose a novel online
trajectory planner by structuring the motion planning problem as sequences
of spatial-temporal graphs passing through a sequential graph neural
network architecture.
\begin{table}[H]
\caption{\textcolor{black}{Nomenclature}\label{tab:Nomenclature}}

\centering{}%
\begin{tabular}{|c|>{\centering}p{0.825in}|c|>{\centering}p{0.825in}|}
\hline 
\textbf{\scriptsize{}Symbol} & \textbf{\scriptsize{}Description} & \textbf{\scriptsize{}Symbol} & \textbf{\scriptsize{}Description}\tabularnewline
\hline 
{\scriptsize{}$s,d$} & {\scriptsize{}Longitudinal, lateral positions} & {\scriptsize{}$V_{s},V_{d}$} & {\scriptsize{}Longitudinal, lateral virtual nodes}\tabularnewline
\hline 
{\scriptsize{}$N_{V}$} & {\scriptsize{}Number of virtual nodes} & {\scriptsize{}$a,\ddot{s},\ddot{d}$} & {\scriptsize{}Acceleration and/or deceleration}\tabularnewline
\hline 
{\scriptsize{}$v,\dot{s},\dot{d}$} & {\scriptsize{}Velocity} & {\scriptsize{}$J$} & {\scriptsize{}Jerk}\tabularnewline
\hline 
{\scriptsize{}$U_{o},U_{v}$} & {\scriptsize{}Obstacle potential, velocity potential } & {\scriptsize{}$y$} & {\scriptsize{}Output vector of longitudinal and lateral positions}\tabularnewline
\hline 
{\scriptsize{}$E$} & {\scriptsize{}Ego vehicle} & {\scriptsize{}$A$} & {\scriptsize{}Actor vehicle}\tabularnewline
\hline 
{\scriptsize{}$G$} & {\scriptsize{}Graph} & {\scriptsize{}$t_{s}$} & {\scriptsize{}Sampling period}\tabularnewline
\hline 
{\scriptsize{}$N$} & {\scriptsize{}Planning Horizon} & {\scriptsize{}$N_{A}$} & {\scriptsize{}Number of actors}\tabularnewline
\hline 
{\scriptsize{}$h$} & {\scriptsize{}Feature or node embedding} & {\scriptsize{}$W,B$} & {\scriptsize{}Weights, biases}\tabularnewline
\hline 
{\scriptsize{}$p,q$} & {\scriptsize{}Target node, neighboring node} & {\scriptsize{}$\alpha$} & {\scriptsize{}Attention coefficient}\tabularnewline
\hline 
\multicolumn{4}{|c|}{\textbf{\scriptsize{}Subscript/Superscript}}\tabularnewline
\hline 
{\scriptsize{}acc, dec} & {\scriptsize{}Acceleration , Deceleration} & {\scriptsize{}upper, lower} & {\scriptsize{}Upper bound, lower bound}\tabularnewline
\hline 
{\scriptsize{}rec, safe} & {\scriptsize{}Recommended, safety} & {\scriptsize{}long, lat} & {\scriptsize{}Longitudinal, Lateral}\tabularnewline
\hline 
\end{tabular}
\end{table}

Different approaches have been undertaken by researchers to generate
feasible trajectories. Sampling-based planners like the Rapidly-exploring
Random Tree (RRT) has also been extensively tested on automated vehicles
for online path planning \cite{karaman2011sampling} due to to ease
of incorporating user-defined objectives. Another sampling-based technique,
state lattice \cite{gu2013focused,mcnaughton2011motion}, discretizes
the state space in a deterministic manner. Although the spatial-temporal
version of the state lattice allows for planning with dynamic obstacles,
its performance depends on sampling density, making it time-consuming
\cite{huang2019motion}. However, the resulting paths from graph search-based planners and sampling-based planners are usually not continuous
and thus jerky \cite{gonzalez2015review}. Interpolating curve planners
are also popular choices, e.g., clothoid curves \cite{broggi2012autonomous},
polynomial curves \cite{petrov2014modeling}, Bezier curves \cite{han2010bezier},
etc. However, these planners require global waypoints defined and
can be time-consuming when managing obstacles in real-time \cite{gonzalez2015review}.
Frenet trajectory planners are also in the family of interpolating
curve planners to generate optimal trajectory but utilize the Frenet
coordinate frame instead of the Cartesian coordinate frame \cite{werling2010optimal}.
Geisslinger et al. \cite{geisslinger2023ethical} use the Frenet planner
to generate candidate trajectories for an ego and then select the
best candidate based on five different ethical principles in line
with the European Commission (EU). 

Graph-based approaches are also commonly found in the literature.
These approaches encompass spatial or spatial-temporal representations,
whether one-dimensional or hierarchical, like a tree across the feasible
driving area \cite{stahl2019multilayer}. Each node of the graph has
an associated cost, and the graph-based algorithms seek to identify
the path, minimizing the cost between the adjacent nodes. Graph search-based planners A{*} \cite{kammel2008team}, hybrid A{*} \cite{montemerlo2008junior},
and variations of these techniques have been widely used. The 2007
DARPA Urban Challenge was won by the vehicle called Boss that utilized
the Anytime D{*} algorithm \cite{ferguson2008motion}. In more recent
work, Han et al. \cite{han2023efficient} used hybrid A{*} to find
collision-free paths and further optimized the path using kinematic
constraints. Many use one-layered graphs in the spatial dimensions
with lateral targets along the road \cite{li2017development,hu2018dynamic}.
Gu et al. \cite{gu2013focused} apply multiple layered graphs with
linear edges along the road. The adjacent nodes, which result in the least
cost, are then used for path optimization. McNaughton et al. \cite{mcnaughton2011motion}
explore both spatial and temporal dimensions in real-time in the search
of a minimal cost path in the graph. A multilayered graph-based trajectory
planner is also proposed by \cite{stahl2019multilayer}, where the
planning task for a racing environment is divided into offline and
online components. The offline part creates multiple drivable trajectories
by connecting nodes in the graph, and then, in real-time, the online
part picks the least expensive global state in the scene. 

Artificial potential field techniques allow us to define potential fields
using potential functions (PFs) for obstacles, road structures and
goals and then plan paths by moving in the descent direction of the
field \cite{rasekhipour2016potential}. In \cite{noto2012steering},
Noto et al. generate the reference path to satisfy the dynamic constraints
and to move the vehicle in the descent direction of the PFs. Rasekhipour
et al. \cite{rasekhipour2016potential} combine the power of model
predictive controller to address dynamic constraints and, the power
of PFs to address obstacles and road structures to generate trajectories
for the ego. An online motion planner for vehicle-like robots proposed
by Chen et al. \cite{chen2022online} uses PFs to generate the initial
path meeting road constraints. The path is then optimized as an unconstrained
weighted objective function for curvature maintenance, obstacle avoidance,
and speed profile. In our work, we propose personalized PFs for obstacle
avoidance and maximum velocity, keeping with priority given to the
former. 

The usage of machine learning for motion planning also has its fair
share in the literature. Sung et al. \cite{sung2021training} use
neural networks to learn a planner online from data created using
off-the-shelf path planning algorithms. The results show that the
paths generated by the neural networks were smoother in contrast to
the original paths. Prediction and planning are combined in \cite{huang2023differentiable}
which deploys a neural network to predict the future states of the surrounding
vehicles (actors) and an initial strategy. These are then forwarded
into an optimization-based differentiable motion planner to determine
the final plan. Yang et al. \cite{yang2022hybrid} use a hybrid approach
where the behavioural learning is done using deep reinforcement learning
(DRL) and the planning is done using a Frenet planner. In \cite{feher2019hybrid},
A Deep Deterministic Policy Gradient (DDPG) planner is presented to
determine the optimal trajectory based on pre-defined initial and
final states and dynamic constraints. However, no obstacle is considered
in the environment, and it requires 40,000 iterations before 
a good-quality trajectory can be achieved. Hoel et al. \cite{hoel2019combining} extend
the AlphaGo Zero algorithm to a continuous state space domain without
self-play and combined planning and learning for tactical decision-making in autonomous driving. \cite{krasowski2020safe} proposes a
safety layer in its reinforcement learning (RL) framework that pilots
the exploration process by limiting actions to the safe subspace of
the whole action space. These safe actions are determined by taking
into account all the possible trajectories of the traffic participants.

In this article, we propose spatial-temporal graphs that incorporate
virtual nodes positioned along the road. These virtual nodes are designed
to meet the road boundary conditions as well as the kinematic constraints
of the ego. The ego and its surrounding actors also are nodes in the
graph. The graphs are processed through a network architecture containing
sub-networks in series that generate the future plan for the ego.\textcolor{black}{{}
In contrast to most prior methods, this future plan can include one
or more of the following driving behaviors: lane-keeping, lane-changing,
car-following, and speed-keeping.} Furthermore, a simple behavioral
layer \textcolor{black}{is introduced to command the kinematic constraints}
of the ego for different driving tasks. The trajectory planner is
tested for different driving tasks, and the results obtained demonstrate
a better performance trade-off compared to the two baselines used in this
work. 

The main contributions of this work are summarized as follows:
\begin{enumerate}
\item We propose a novel spatial-temporal graph that depicts the trajectory
planning problem and incorporates road constraints and kinematic constraints.
\item We present a neural network architecture that can process the above-mentioned
graph to generate future plan for the ego.
\item We introduce personalized PFs for the architecture to learn on.
\item We design a simple behavioral layer for \textcolor{black}{defining
kinematic constraints} of the ego.
\end{enumerate}
The rest of this article is organized as follows. Section \ref{sec:Preliminaries}
provides brief descriptions of the theoretical framework used in this
work. The proposed Spatial-Temporal Graph (STG) Trajectory Planner
is elaborated in Section \ref{sec:Spatial-Temporal-Graph-(STG)}.
In Section \ref{sec:Experimental-Results-and}, we discuss the experimental
setup and the results. Finally, Section \ref{sec:Conclusion} concludes
the article.

\section{Theoretical Background\label{sec:Preliminaries}}

\subsection{Graph Neural Network\label{subsec:Graph-Neural-Network}}

Graphs are used as a means of illustrating real-world problems. More
particularly, graphs can capture the interactions between agents and
how these agents evolve because of their involvement in a neighborhood.
The agents in a graph are called the nodes and the interactions between
them are presented using edges. Graphs can be of two types: homogeneous
and heterogeneous. Heterogeneous graphs, unlike homogeneous graphs,
have nodes and/or edges that can have various types or labels associated
with them, indicating their different roles or semantics. For example,
in our work, the node representing the ego has four features while
the nodes for the surrounding actors have two features requiring a
heterogeneous graph representation to capture the interactions between
them. Because of the differences in type and dimensionality, a single
node or edge feature tensor is unable to accommodate all the node
or edge features of the graph \cite{Heterogeneous_PyG}. The computation
of messages and update functions is conditioned on node or edge type.
There are PyTorch libraries that are available to process heterogeneous
graphs, which are detailed in \cite{Heterogeneous_PyG}.

Neural networks that operate on graphs are called Graph Neural Networks
(GNNs) \cite{scarselli2008graph}. For example, in node representations,
the GNN maps nodes of a homogeneous graph to a matrix $\mathbb{R}^{n\times m}$,
where, $n$ is the number of nodes and $m$ is the dimensionality
of the features of the nodes. For node $p$, the GNN aggregates information
from its neighbors (e.g. averages the messages from its neighbors)
and then applies a neural network in several layers \cite{stanfordGNN}:
\begin{equation}
h_{p}^{(i+1)}=\sigma\left(W_{i}\underset{q\in\mathcal{N}(p)}{\sum}\frac{h_{q}^{(i)}}{\left|\mathcal{N}(p)\right|}+B_{i}h_{p}^{(i)}\right),
\end{equation}
$\forall i\in\left\{ 0,1,\ldots,L-1\right\} $, where, $h$ is the
embedding of a node, $\sigma$ is a non-linear activation function,
$W_{i}$ and $B_{i}$ are the weights and biases of the $i^{\mathrm{th}}$
layer, respectively, $\mathcal{N}(p)$ is the neighborhood of the
target node $p$, and $L$ is the total number of layers.

Graph Attention Network (GAT) \cite{velivckovic2017graph} is a GNN
that integrates attention so that the learning is focused on more
relevant segments of the input. The network learns the importance
(also called the attention coefficient $e_{pq}$) of the neighbors
of a node as it aggregates information from them. The attention coefficient
between the target node $p$ and its neighbor $q$ is computed by
applying a common linear transformation $\boldsymbol{W}$ to the features
($\boldsymbol{h}$) of both $p$ and $q$, followed by a shared attentional
mechanism ($\mathrm{att}$) as follows:
\begin{equation}
e_{pq}=\mathrm{att}\left(\boldsymbol{W}\boldsymbol{h}_{p},\boldsymbol{W}\boldsymbol{h}_{q}\right)
\end{equation}
A single-layered feed-forward neural network is used for $\mathrm{att}$
in \cite{velivckovic2017graph}. To compare the neighboring nodes
properly, $e_{pq}$ is normalized using the softmax function:
\begin{equation}
\alpha_{pq}=\frac{\mathrm{exp}\left(e_{pq}\right)}{\sum_{k\in\mathcal{N}(p)}\mathrm{exp}\left(e_{pk}\right)}\label{eq:attention_coeff}
\end{equation}
For a detailed explanation of GAT, readers can refer to \cite{velivckovic2017graph}. 

\subsection{Frenet Coordinates}

\begin{figure}[t]
\centering{}\subfloat[\label{fig:frenet_frame_2}]{\begin{centering}
\includegraphics[scale=0.22]{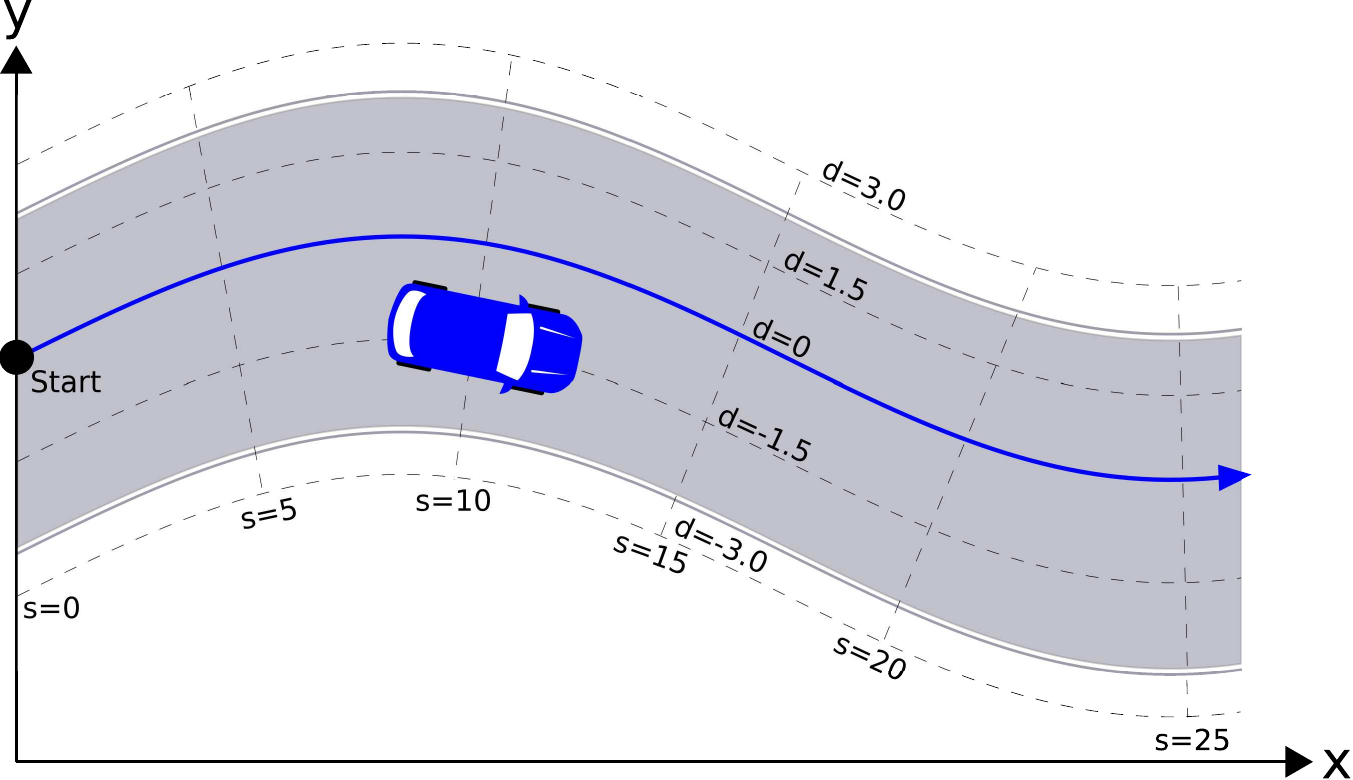}
\par\end{centering}
}\hspace{0.1in}\subfloat[\label{fig:Frenet-frame_1}]{\begin{centering}
\includegraphics[scale=0.22]{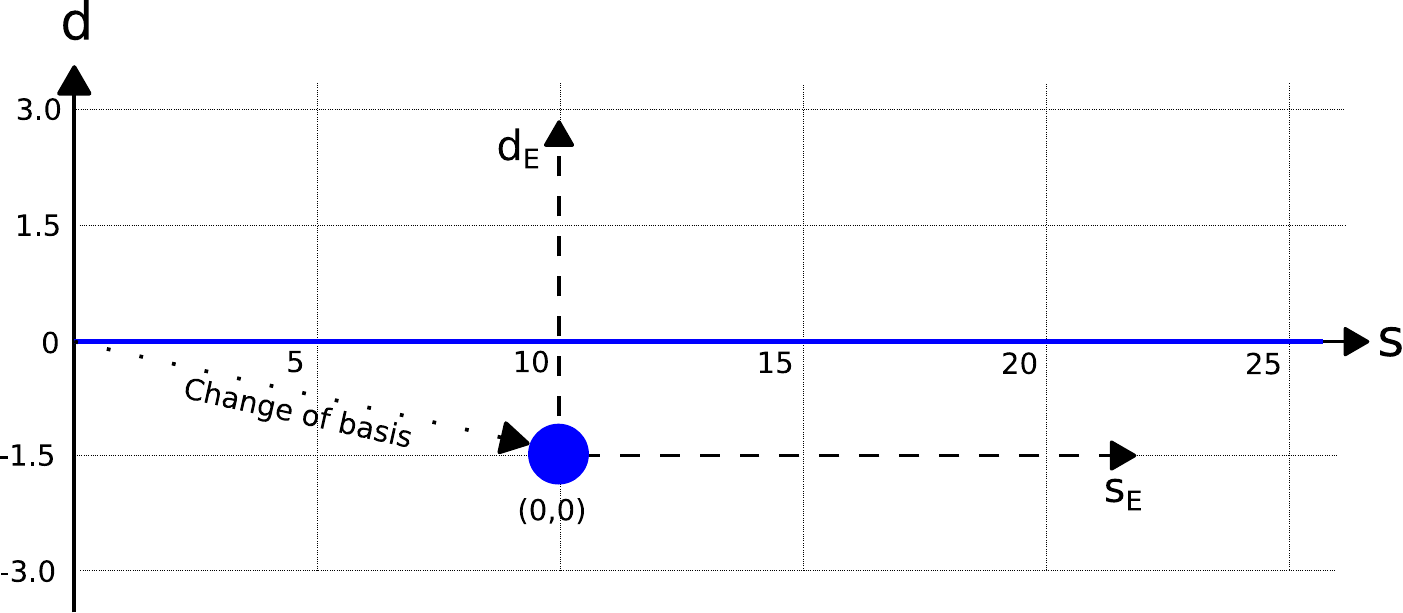}
\par\end{centering}
}\caption{(a) A snapshot of the road presented in the Cartesian coordinate frame
showing the Frenet framework formulation with respect to the reference
curve (in blue), and, (b) Frenet coordinate frame of the snapshot
with the ego (blue circle). The new Frenet coordinate frame (black
solid dashes) is with respect to the ego position}
\end{figure}

Frenet coordinate system \cite{werling2010optimal} is used to describe
the location of a point relative to a reference curve. More particularly,
the Frenet coordinates $s$ and $d$ represent the longitudinal and
the lateral distances of the point of interest with respect to the
curve from a starting point. Figure \ref{fig:frenet_frame_2} shows
the Frenet coordinates inside a Cartesian coordinate frame with respect
to a reference curve.

It is important to note that any road structure (curved, straight,
etc.) in the Cartesian coordinate framework can be presented as a
straight line in the Frenet coordinate framework, as shown in Figure
\ref{fig:Frenet-frame_1}. Thus, presenting an autonomous driving
task in the latter domain is much simpler. In this work, we use the
Frenet coordinates as features. Furthermore, for this work, after
the problem is translated in the Frenet domain, the basis of the new
frame is shifted $\left(\left\{ s,d\right\} \rightarrow\left\{ s_{E},d_{E}\right\} \right)$
with respect to the ego as shown in Figure \ref{fig:Frenet-frame_1}.
By applying this shift, it can be ensured that the magnitudes of the
virtual nodes, as will be discussed in section \ref{subsec:Spatial-Temporal-Graphs},
always remain in a constrained range. Lastly, the Frenet coordinate
frame also allows us to define the cost of a path very effectively
based on the longitudinal and the lateral displacements of the ego
and the other actors.
\begin{figure}[t]
\begin{centering}
\includegraphics[scale=0.11]{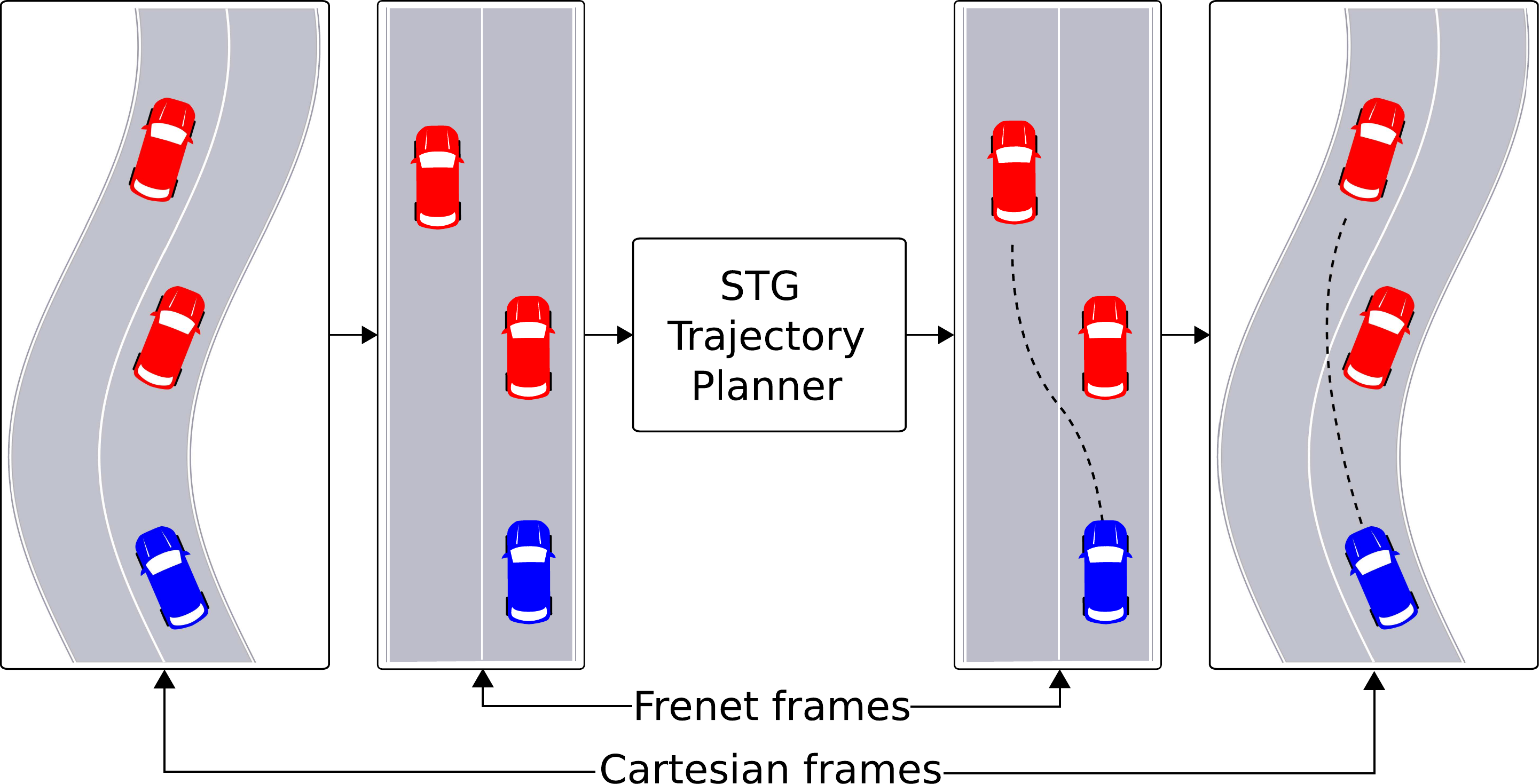}
\par\end{centering}
\caption{\textcolor{black}{Transformations between the Cartesian and the Frenet
frames to obtain trajectory (in dashed line)}\label{fig:frame_conversion}}
\end{figure}

\textcolor{black}{In this work, a scenario given in the Cartesian
frame is first transformed into the Frenet frame, which is used to solve
the trajectory problem using our proposed planner. The trajectory
obtained in the Frenet frame is then transformed back to the Cartesian
frame to get the actual trajectory for the ego. These steps are illustrated
in Figure \ref{fig:frame_conversion}.}
\begin{figure*}[t]
\begin{centering}
\includegraphics[scale=0.19]{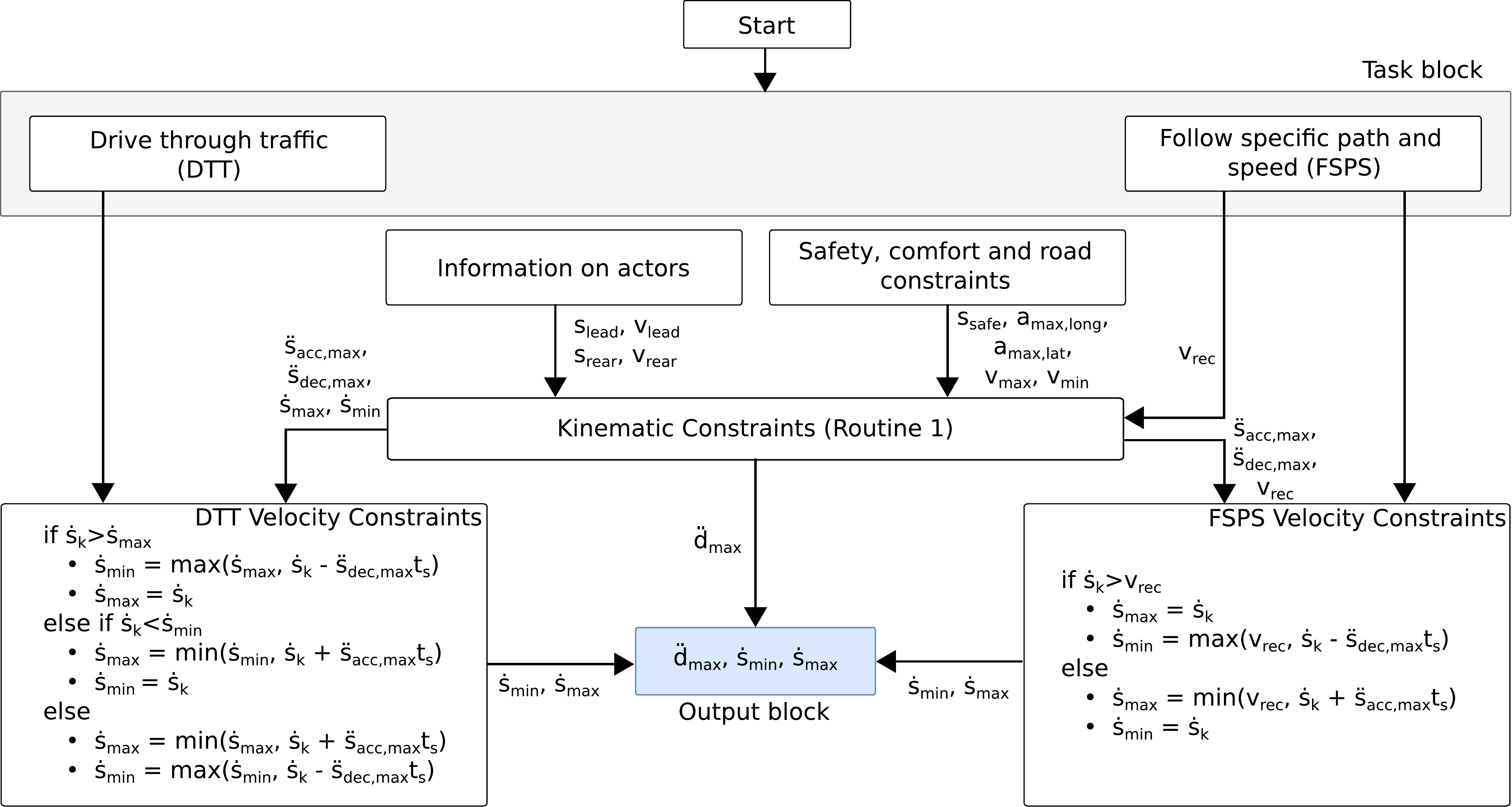}
\par\end{centering}
\caption{Flow diagram for the behavioral layer (see Appendix for Routine 1)
\label{fig:behavioral layer}}
\end{figure*}

\section{Spatial-Temporal Graph (STG) Trajectory Planner\label{sec:Spatial-Temporal-Graph-(STG)}}

In this section, the GNN-based trajectory planner utilizing a sequence
of spatial-temporal graphs for trajectory planning is presented. \textcolor{black}{For
the formation of the planner, first, we discuss a simple behavioral
layer that determines the kinematic constraints for the ego. Next,
the formulation of the spatial graphs considering the road constraints,
the kinematic constraints of the ego and the actors surrounding it
are elaborated. }Then, we present the network architecture that processes
these graphs to obtain the future trajectory. Lastly, we define the
cost function to be minimized by the network for safety. Even though
our work is on the design of the trajectory planner module, there
are other vital modules in an autonomous vehicle system. Since the
design of those other modules is out of the scope of this work, it is
assumed that the states of the actors, road structure and the regulations
are received from the perception module and the future states of the
actors are received from the prediction module.

\subsection{Behavioral Layer\label{subsec:Behavioral-Layer}}

\textcolor{black}{In this section, we present a simple behavioral
layer for our proposed framework -- the design of a more sophisticated
behavioral layer is out of the scope of this work. The behavioral layer
determines kinematic constraints for the ego for different driving
tasks, considering safety with respect to actors, comfort, and, road
regulations. A kinematic constraint refers to a limitation or condition
imposed on the vehicle's motion based on its physical characteristics
and the surrounding environment.}

\textcolor{black}{At first, parameters prioritizing safety (safety
gap $s_{\mathrm{safe}}$), comfort (the maximum longitudinal acceleration/deceleration
$a_{\mathrm{max,long}}$, the maximum lateral acceleration/deceleration
$a_{\mathrm{max,lat}}$) and road regulations (recommended speed $v_{\mathrm{rec}}$
if any, maximum road speed $v_{\mathrm{max}}$, minimum road speed
$v_{\mathrm{min}}$) are defined. Next, the longitudinal gap with rear
($s_{\mathrm{rear}}$) and/or lead ($s_{\mathrm{lead}}$) actor(s),
and its velocity ($v_{\mathrm{rear}}$ and/or $v_{\mathrm{lead}}$)
are identified. Note that a lead or a rear actor refers to actors
which are immediate and are in the same lane as the ego. Using
these parameters, the longitudinal and lateral kinematic constraints
$\ddot{s}_{\mathrm{dec,max}}$, $\ddot{s}_{\mathrm{acc,max}}$, $\dot{s}_{\mathrm{max}}$,
$\dot{s}_{\mathrm{min}}$, $\ddot{d}_{\mathrm{max}}$ and $v_{\mathrm{rec}}$
(see Table \ref{tab:Nomenclature} for description) are obtained from
Routine \ref{alg:Kinematic Constraints} (see Appendix). The behavioral
layer, presented as a flow diagram in Figure \ref{fig:behavioral layer},
is designed to address two particular driving tasks: driving through
traffic (DTT), and following a specific path and speed (FSPS). Using
the updated constraints, the behavioral layer generates the longitudinal
velocity constraints ($\dot{s}_{\mathrm{max}}$, $\dot{s}_{\mathrm{min}}$)
and the lateral acceleration constraint ($\ddot{d}_{\mathrm{max}}$
) required in the formulation of a spatial-temporal graph which will
be discussed in the next subsection.}

\subsection{Spatial-Temporal Graphs\label{subsec:Spatial-Temporal-Graphs}}

A self-driving vehicle must respect the road constraints, e.g., road/lane
boundaries, speed limit, etc., to ensure safety. In addition, it cannot
violate the kinematic constraints, e.g., steering limits, acceleration/deceleration
limits. In our proposed formulation of the graph, we aim to incorporate
some of these constraints. Note that the formation of the graphs requires
knowledge of the future trajectories of the surrounding actors.

\begin{figure}[t]
\begin{centering}
\includegraphics[scale=0.5]{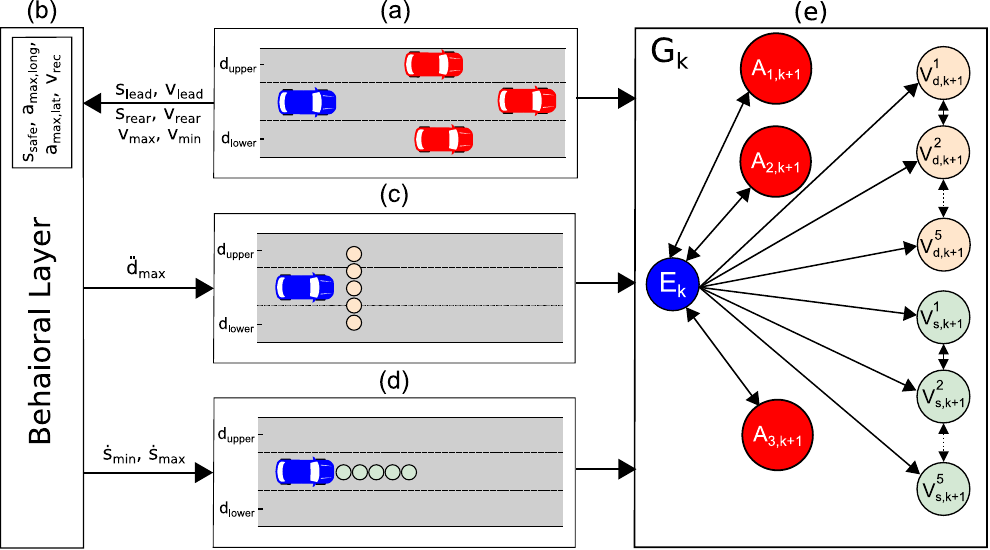}
\par\end{centering}
\caption{(a) The ego (in blue) is surrounded by actors (in red).\textcolor{red}{{}
}\textcolor{black}{(b) The behavioral layer generates kinematic constraints
for the current scenario. }(c) The lateral virtual nodes (in light
peach) is spread out laterally along the road, respecting road boundaries.
(d) The longitudinal virtual nodes (in light green) are spread out
longitudinally ahead of the ego along the road. (e) Formation of graph
$G_{k}$ for the given snapshot with $N_{V}=5$ \label{fig:Formation-of-graph}}
\end{figure}

An example snapshot is depicted in Figure \ref{fig:Formation-of-graph}(a).
In this snapshot, the ego is surrounded by three other actors. The
ego is presented as the node $E_{k}$ in the graph with its longitudinal
and lateral positions and velocities as features at the $k^{\mathrm{th}}$
step with $k=0,\ldots,N-1$, where $N$ is the planning horizon and
$k=0$ denotes the initial timestamp. All of the $N_{A}$ number of
actors in a scenario are represented by nodes $A_{i,k+1}$ with $i=1,\ldots,N_{A}$.
The future lateral and longitudinal positions of the actors at the
$\left(k+1\right)^{\mathrm{th}}$ step are the features of their corresponding
nodes. Next, we introduce the idea of virtual nodes $V$ -- these
are imaginary points on the road spaced laterally or longitudinally
-- using kinematic constraints received from the behavioral layer
as shown in Figure \ref{fig:Formation-of-graph}(b) -- with respect
to the ego as shown in Figure \ref{fig:Formation-of-graph}(c) and
Figure \ref{fig:Formation-of-graph}(d), respectively. Thus, these
virtualized nodes have only one feature -- the lateral or the longitudinal
spacing from the ego at $k^{\mathrm{th}}$ step. Note that the lateral
and the longitudinal positions discussed are with respect to a reference
path. In the remainder of the paper, we will use Frenet coordinates
$d$ and $s$ for lateral displacement and longitudinal distances,
respectively.

\textcolor{black}{The lateral nodes are constrained by the road/lane
boundaries and the lateral kinematic constraint $\ddot{d}_{\mathrm{max}}$
obtained from the behavioral layer. At the $\left(k+1\right)^{\mathrm{th}}$
step, the maximum lateral velocity of the ego is given by}
\begin{equation}
\dot{d}_{\mathrm{max}}^{k+1}=\ddot{d}_{\mathrm{max}}t_{s}\label{eq:max_lat_vel}
\end{equation}
where, $t_{s}$ is the sampling period. Therefore, equation (\ref{eq:max_lat_vel})
imposes the lateral kinematic constraints. The road constraint, i.e.,
$d_{\mathrm{lower}}\leq d^{k+1}\leq d_{\mathrm{upper}}$, where the
boundary conditions $d_{\mathrm{lower}}$ and $d_{\mathrm{upper}}$
constrain the ego to stay on the road, is imposed using the following:
\begin{equation}
\begin{aligned}d_{\mathrm{max}}^{k+1} & =\mathrm{min}\left(d_{\mathrm{upper}},d^{k}+\dot{d}_{\mathrm{max}}^{k+1}t_{s}\right)\\
d_{\mathrm{min}}^{k+1} & =\mathrm{max}\left(d_{\mathrm{lower}},d^{k}-\dot{d}_{\mathrm{max}}^{k+1}t_{s}\right)
\end{aligned}
\label{eq:lat_d}
\end{equation}
where, $d_{\mathrm{max}}^{k+1}$ and $d_{\mathrm{min}}^{k+1}$ are
the maximum and the minimum lateral distances that the ego can move.
Given the number of lateral virtual nodes $N_{V}$, the nodes are
layered laterally with equal spacing as follows:
\begin{equation}
\begin{aligned}V_{d,k+1}= & \left[\begin{array}{cccc}
d_{\mathrm{min}}^{k+1} & d_{\mathrm{min}}^{k+1}+\delta_{d} & d_{\mathrm{min}}^{k+1}+2\delta_{d} & \cdots\end{array}\right.\\
 & \left.\begin{array}{cc}
d_{\mathrm{max}}^{k+1}-\delta_{d} & d_{\mathrm{max}}^{k+1}\end{array}\right]^{T}\in\mathbb{R}^{N_{V}}
\end{aligned}
\label{eq:Vd_spaced}
\end{equation}
where, $\delta_{d}$ is the spacing defined by
\begin{equation}
\delta_{d}:=\frac{\left(d_{\mathrm{max}}^{k+1}-d_{\mathrm{min}}^{k+1}\right)}{\left(N_{V}-1\right)}.\label{eq:del_d}
\end{equation}

\textcolor{black}{For the longitudinal nodes, the kinematic constraints
$\dot{s}_{\mathrm{max}}$ and $\dot{s}_{\mathrm{min}}$ obtained from
the behavioral layer is applied. }Thus, the maximum and the minimum
longitudinal distances the ego can traverse are defined by
\begin{equation}
\begin{aligned}s_{\mathrm{max}}^{k+1} & =s^{k}+\dot{s}_{\mathrm{max}}t_{s}\\
s_{\mathrm{min}}^{k+1} & =s^{k}+\dot{s}_{\mathrm{min}}t_{s}
\end{aligned}
,\label{eq:long_pos}
\end{equation}
respectively. For $N_{V}$ number of virtual longitudinal nodes, the
longitudinal layering of these nodes are defined similarly as above:
\begin{equation}
\begin{aligned}V_{s,k+1}:= & \left[\begin{array}{cccc}
s_{\mathrm{min}}^{k+1} & s_{\mathrm{min}}^{k+1}+\delta_{s} & s_{\mathrm{min}}^{k+1}+2\delta_{s} & \cdots\end{array}\right.\\
 & \left.\begin{array}{cc}
s_{\mathrm{max}}^{k+1}-\delta_{s} & s_{\mathrm{max}}^{k+1}\end{array}\right]^{T}\in\mathbb{R}^{N_{V}}
\end{aligned}
\label{eq:Vs_spaced}
\end{equation}
where, $\delta_{s}$ is the equal spacing between the adjacent nodes
defined by
\begin{equation}
\delta_{s}:=\frac{\left(s_{\mathrm{max}}^{k+1}-s_{\mathrm{min}}^{k+1}\right)}{\left(N_{V}-1\right)}\label{eq:del_s}
\end{equation}
After all the nodes ($E_{k}$, $A_{i,k+1}$, $V_{s,k+1}$ and $V_{d,k+1}$)
have been defined, the graph $G_{k}$ is formed as shown in Figure
\ref{fig:Formation-of-graph}(e). The edges between $E_{k}$ and $A_{i,k+1}$
are bidirectional signifying the ``\textit{interactions}'' between
the ego and the other actors -- the edge features are the euclidean
distances between the ego and the corresponding actors. The edges
between the ego and the virtual nodes are unidirectional ($E$ to
$V_{s}$ and $V_{d}$) signifying the transition from $k^{\mathrm{th}}$
to $\left(k+1\right)^{\mathrm{th}}$ step (the temporal aspect of
the graph itself). Thus, the edge attribute for the edges between
$E$ and, $V_{s}$ and $V_{d}$, is set to be the sampling period
$t_{s}$. The lateral virtual nodes are connected by bidirectional
edges to their adjacent nodes with each edge attribute being the corresponding
the lateral distance between the corresponding nodes. The interactions
between the longitudinal nodes are formulated in the same manner.

\subsection{STG Network Architecture}

\begin{figure}[h]
\begin{centering}
\includegraphics[scale=0.3]{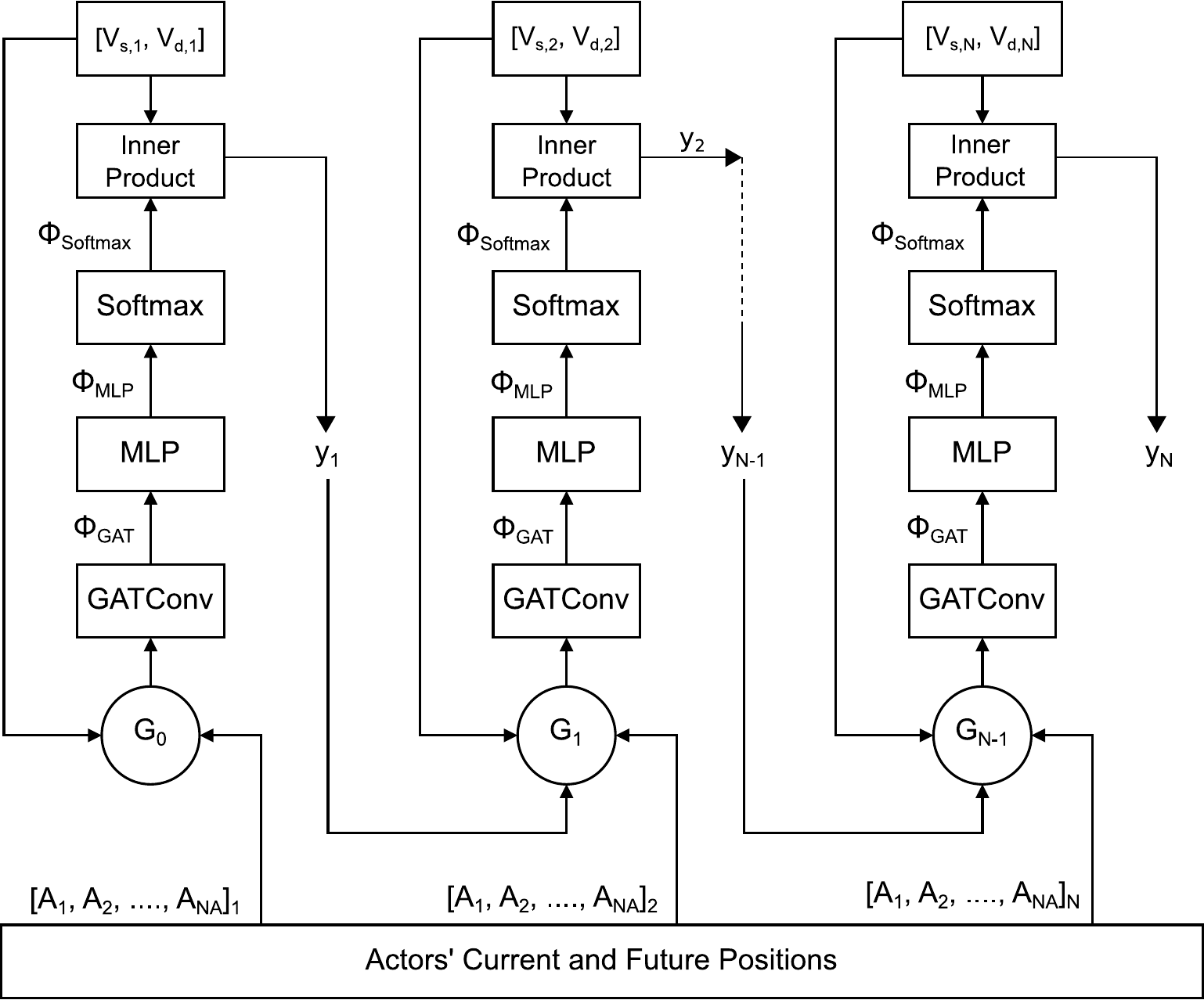}
\par\end{centering}
\caption{Network architecture to learn online the future trajectory of the
ego \label{fig:Network-architecture}}
\end{figure}

\textcolor{black}{The task of the STG network architecture is to output
the future longitudinal and lateral positions of the ego. To do that,
we utilize a GAT network \cite{velivckovic2017graph} as an encoder
to generate a node embedding of the spatial-temporal graph $G_{k}$
(equation (\ref{eq:GATConv})) and create the encoded feature vector
$\Phi_{\mathrm{GAT}}$ (equations (\ref{eq:sum_actors}) and (\ref{eq:GAT_output})).
A multilayer perceptron (MLP) decoder with a fixed-size input further
processes $\Phi_{\mathrm{GAT}}$ to generate the decoded features
$\Phi_{MLP}$ (equation (\ref{eq:MLP_output})). A softmax function
is then applied on $\Phi_{MLP}$ in two layers to compute two sets
of weights $\Phi_{\mathrm{Softmax}}^{s}$ and $\Phi_{\mathrm{Softmax}}^{d}$
for the longitudinal and lateral virtual nodes, respectively (equation
(\ref{eq:softmax})). These weights determine the relative importance
of the virtual nodes. Finally, inner products between the virtual
nodes and the weights are computed to obtain the future position $y_{k+1}$
of the ego at the $\left(k+1\right)^{\mathrm{th}}$ step (equation
(\ref{eq:inner_product})). Each of the mentioned sequential steps
are further detailed below and shown in Figure \ref{fig:Network-architecture}.}

The graph $G_{k}$ that represents the snapshot of the ego's surrounding
at the $k^{\mathrm{th}}$ step needs to be processed to identify the
position of the ego at the $\left(k+1\right)^{\mathrm{th}}$ step.
In order to do so, we propose the framework shown in Figure \ref{fig:Network-architecture}.
At first, $G_{k}$ is passed through a GAT network that generates\textcolor{black}{{}
an embedding of size $r$ for each of the nodes, i.e.,
\begin{equation}
\begin{aligned}\left[\Phi_{\mathrm{GAT}}^{E},\Phi_{\mathrm{GAT}}^{V_{s}},\Phi_{\mathrm{GAT}}^{V_{d}},\Phi_{\mathrm{GAT}}^{A_{1}},\ldots,\Phi_{\mathrm{GAT}}^{A_{N_{A}}}\right]=\\
\mathrm{GATConv}\left(G_{k}\right)
\end{aligned}
\label{eq:GATConv}
\end{equation}
where, $\Phi_{\mathrm{GAT}}^{E}\in\mathbb{R}^{r}$ for the ego, $\Phi_{\mathrm{GAT}}^{V_{s}}\in\mathbb{R}^{N_{V}\times r}$
for the $N_{V}$ longitudinal virtual nodes, $\Phi_{\mathrm{GAT}}^{V_{d}}\in\mathbb{R}^{N_{V}\times r}$
for the $N_{V}$ lateral virtual nodes, and $\Phi_{\mathrm{GAT}}^{A_{1}},\Phi_{\mathrm{GAT}}^{A_{2}},\ldots,\Phi_{\mathrm{GAT}}^{A_{N_{A}}}\in\mathbb{R}^{r}$
for the $N_{A}$ actors.} Note that the varying number of actors $N_{A}$
surrounding the ego cause $G_{k}$ to be a dynamic graph, i.e., addition
and deletion of actor nodes from $G_{k}$. 

\textcolor{black}{Since MLPs expect a fixed-size input and the GAT
network applied to $G_{k}$ may have different sizes depending on
$N_{A}$, we address this structural problem by applying graph pooling
\cite{hamilton2020graph} on the actor nodes only:
\begin{equation}
\Phi_{\mathrm{GAT}}^{A}=\Phi_{\mathrm{GAT}}^{A_{1}}+\Phi_{\mathrm{GAT}}^{A_{2}}+\ldots+\Phi_{\mathrm{GAT}}^{A_{N_{A}}}\label{eq:sum_actors}
\end{equation}
such that $\Phi_{\mathrm{GAT}}^{A}\in\mathbb{R}^{r}$.}\textcolor{blue}{{}
}Graph pooling based on the sum is often adequate for applications
involving small graphs as used in this work \cite{hamilton2020graph}.
Finally, the output of the GAT network is generated by concatenation
as follows:\textcolor{black}{
\begin{equation}
\Phi_{\mathrm{GAT}}=\mathrm{concat}\left(\Phi_{\mathrm{GAT}}^{E},\Phi_{\mathrm{GAT}}^{A},\left(\Phi_{\mathrm{GAT}}^{V_{s}}\right)^{f},\left(\Phi_{\mathrm{GAT}}^{V_{d}}\right)^{f}\right)\label{eq:GAT_output}
\end{equation}
where, $\Phi_{\mathrm{GAT}}\in\mathbb{R}^{2r+2rN_{V}}$ and the $\left(\cdot\right)^{f}$
operator is the flattening function \cite{Flatten_pytorch} that returns
an array collapsed into one dimension.} With $\Phi_{\mathrm{GAT}}$
as the input, the MLP decoder generates the decoded states as follows:\textcolor{black}{
\begin{equation}
\Phi_{\mathrm{MLP}}=\mathrm{MLP}\left(\Phi_{\mathrm{GAT}}\right)\in\mathbb{\mathbb{R}}^{2N_{V}}\label{eq:MLP_output}
\end{equation}
Note that the output size for the MLP decoder is set to $2N_{V}$
since the idea is to create weights for the $N_{V}$ longitudinal
nodes and the $N_{V}$ lateral nodes by applying softmax function
$S$. Then $S$ is used to normalize $\Phi_{\mathrm{MLP}}$ in two
layers (first $N_{V}$ elements of $\Phi_{\mathrm{MLP}}$ for longitudinal
weights and the remaining $N_{V}$ elements for lateral weights) as
follows:}
\begin{equation}
\begin{aligned}\Phi_{\mathrm{Softmax}}^{s} & =S\left(\Phi_{\mathrm{MLP}}^{1:N_{V}}\right)\in\mathbb{\mathbb{R}}^{N_{V}}\\
\Phi_{\mathrm{Softmax}}^{d} & =S\left(\Phi_{\mathrm{MLP}}^{N_{V}+1:2N_{V}}\right)\in\mathbb{\mathbb{R}}^{N_{V}}
\end{aligned}
\label{eq:softmax}
\end{equation}
\textcolor{black}{Finally, the virtual lateral and longitudinal nodes
are weighted using $\Phi_{\mathrm{Softmax}}^{d}$ and $\Phi_{\mathrm{Softmax}}^{s}$
to get the output}
\begin{equation}
y_{k+1}=\left[\begin{array}{cc}
\left\langle \Phi_{\mathrm{Softmax}}^{s},V_{s,k+1}\right\rangle  & \left\langle \Phi_{\mathrm{Softmax}}^{d},V_{d,k+1}\right\rangle \end{array}\right]^{T}\label{eq:inner_product}
\end{equation}
where, \textcolor{black}{$y_{k+1}=\left[\begin{array}{cc}
s^{k+1} & d^{k+1}\end{array}\right]^{T}\in\mathbb{R}^{2}$ is a vector of the future longitudinal and lateral positions at the
$\left(k+1\right)^{\mathrm{th}}$ step}, and, the operator $\left\langle \right\rangle $
represents the inner product. $y_{k+1}$ is then utilized to obtain
$G_{k+1}$ and the same routine is followed to obtain $y_{k+2}$ and
so on as shown in Figure \ref{fig:Network-architecture}.

\subsection{Potential Functions}

The network needs to backpropagate some ``\textit{loss}'' to learn.
However, in an online trajectory planner, there is no labelled trajectory,
and thus, no loss can be calculated. Therefore, we propose potential
functions (PFs) of the obstacles that determine the safety of the
ego for a given trajectory. The obstacle PF has a maximum value in its
position to repel the ego.

\begin{figure}[t]
\begin{centering}
\includegraphics[scale=0.23]{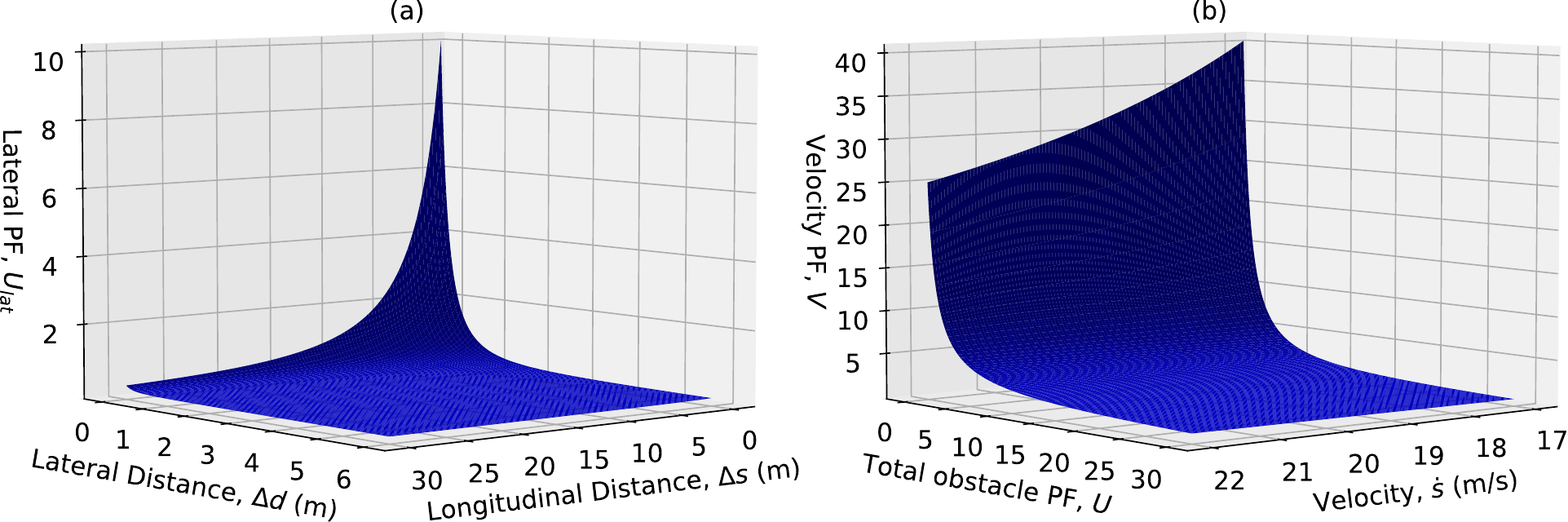}
\par\end{centering}
\centering{}\caption{Potential functions -- (a) Lateral potential function, and, (b) Velocity
potential function \label{fig:Penalty-functions}}
\end{figure}

For an actor longitudinally and laterally distanced from the ego by
$\triangle s$ and $\triangle d$, respectively, is assigned the longitudinal
and the lateral potential functions as follows:
\begin{equation}
U^{\mathrm{long}}:=\frac{b_{1}}{\left(b_{2}\triangle s+\varepsilon_{1}\right)^{2}}\label{eq:risk_long}
\end{equation}
\begin{equation}
U^{\mathrm{lat}}:=\frac{b_{3}U^{\mathrm{long}}}{\left(b_{4}\triangle d+\varepsilon_{1}\right)^{2}}\label{eq:risk_lat}
\end{equation}
respectively, where $b_{1},$ $b_{2},$ $b_{3},$ $b_{4},$ and $\varepsilon_{1}$
are constants. \textcolor{black}{Equation (\ref{eq:risk_long}) --
a slight variation of the repulsive potential proposed by \cite{khatib1985real}
}-- simply indicates that maintaining a larger longitudinal distance
alone is safer. However, safety with respect to lateral distance alone
is not sufficient. An actor maintaining a particular lateral distance
from the ego at a very close proximity in terms of the longitudinal
distance possesses higher risk than an actor maintaining the same lateral
distance but further away longitudinally. \textcolor{black}{Thus,
we propose equation (\ref{eq:risk_lat}) }that accounts for this phenomenon
and Figure \ref{fig:Penalty-functions}(a) illustrates it. For $N_{A}$
number of actors surrounding the ego, the \textcolor{black}{total
potential function of the obstacles $U_{o}$ }is determined as follows:\textcolor{black}{
\begin{equation}
U_{o}:=\stackrel[i=1]{N_{A}}{\sum}U_{i}^{\mathrm{lat}}\label{eq:risk_score}
\end{equation}
}Next, it is important to realize that when the ego is driven through
traffic, the planner network may slow the ego so that the surrounding
actors can pass by suggesting a lower potential $U_{o}$. However,
this is an ineffective way of solving the trajectory plan. Thus,
to address the issue,\textcolor{black}{{} a potential function of velocity
$U_{v}$ is also introduced }in this work:\textcolor{black}{
\begin{equation}
U_{v}:=c_{1}\left(\frac{c_{2}}{U_{o}+\varepsilon_{2}}\right)^{\tfrac{\dot{s}_{\mathrm{max}}}{\dot{s}}}\label{eq:velocity_pot}
\end{equation}
}where $c_{1},$ $c_{2},$ and $\varepsilon_{2}$ are constants. Equation
(\ref{eq:velocity_pot}) implies that lower velocities should have
higher potential when the risk is lower, i.e., $U_{o}$ is lower.
Thus, the velocity PF has its maximal value when the risk is the lowest
and the ego is travelling at the minimum velocity. On the other hand,
when $U_{o}$ is higher, the velocity PF is minimal since safety is
a higher priority. Figure \ref{fig:Penalty-functions}(b) illustrates
$U_{v}$\textcolor{black}{. Finally, the total potential $U_{\mathrm{total}}$
(i.e., the ``loss'')} backpropagated into the network is given by\textcolor{black}{
\begin{equation}
U_{\mathrm{total}}=\stackrel[k=0]{N-1}{\sum}\left(U_{o}^{k+1}+U_{v}^{k+1}\right)\label{eq:cost}
\end{equation}
}

\section{Experimental Results and Analysis\label{sec:Experimental-Results-and}}

\subsection{Driving Tasks}

In this section, some autonomous driving tasks are defined to evaluate
the performance of the proposed planner. These tasks vary in complexity,
e.g., vehicle following and lane keeping are on the simpler side of
the spectrum, whereas, merging, and driving through traffic are at the
other end \cite{aradi2020survey}. Often, simpler tasks are sub-tasks
of more complex tasks. Particularly, we define three complex problem
statements -- vehicle following and lane keeping are sub-tasks of
the defined problems. 

\subsubsection{Driving Through Traffic}

Aradi \cite{aradi2020survey} suggests that driving through traffic
is the most complicated setup to test a planner. Although this task
is scalable, we limit the task to highway driving. Three different
densities -- low, medium, and high -- of traffic scenarios (100
for each) were generated using the SL2015 model of SUMO (Simulation
of Urban MObility) simulator \cite{lopez2018microscopic}. The low-density traffic scenario consists of 1-5 actors, the medium-density
traffic scenarios consist of 10-14 actors surrounding the ego, and
the high ones consist of 15-20 actors surrounding the ego. The actors
are set to respect the speed limits of the road. Once the traffic
is generated, an agent from the scene is randomly selected to play
the role of the ego while the trajectories of other agents remain
unchanged. In this task, the ego has to plan a safe trajectory for
a 5 seconds horizon with 0.1 second sampling time.

\subsubsection{Merging\label{subsec:Merging}}

In this task, the ego (driving at $\textrm{50 km/h}$) has to merge
to a highway and its lane ends in approximately $\textrm{100 m}$.
The recommended speed to enter the merging lane is set at $v_{\mathrm{rec}}=\textrm{60 km/h}$.
There is an actor approaching with a speed of $\textrm{70 km/h}$
in the merging lane of the highway. There is another actor in the
adjacent lane driving at $\textrm{75 km/h}$.

\subsubsection{Taking an Exit\label{subsec:Taking-an-Exit}}

The ego currently driving at $\textrm{80 km/h}$ has to take an exit
with the recommended speed of $v_{\mathrm{rec}}=\textrm{50 km/h}$.
There is an actor approximately $\textrm{25 m}$ ahead of it driving
at $\textrm{55 km/h}$. In the course of time, the actor also takes
the exit, reducing its speed to the recommended speed. Another actor
is driving in the adjacent lane at $\textrm{75 km/h}$, which does
not take the exit.

\subsection{Baselines}

Two baselines are used in this paper to assess and compare our proposed
technique. The kinematic constraints used in both baselines are
the same used in the proposed method. 

SL2015 from SUMO is used as the first baseline. The SL2015 model is
a lane-changing behavior model used to replicate real-world lane-changing
maneuvers in a traffic simulation. The model allows for the definition of kinematic
constraints such as the maximum lateral and longitudinal speeds. The
model makes lane changing decisions based on several factors, such
as the difference between the current and the desired speeds, cooperation
between the driving agents, and the safety gap in the current lane.
The data generated using the model already defines the trajectory
of the randomly chosen agent as the ego. Thus, the trajectory not
only respects the kinematic constraints but also mimics cooperative
behavior with other actors on the road to maintain safety.

Next, we use the Frenet path planner from MATLAB (MFP) \cite{matlabNavigation2022}.
It generates multiple candidate trajectories using fourth or fifth-order
polynomials relative to the reference path. The candidates are then
pruned by checking for kinematic constraints. Next, the trajectories
are assessed for collision against the predicted motions of the surrounding
actors and the colliding ones are eliminated. Finally, the cost of
the remaining trajectories is measured, and the least expensive trajectory
is selected. Equation (\ref{eq:cost}) is used for measuring the cost. 

\subsection{Evaluation Metrics}

The trajectory qualities are evaluated in terms of safety, comfort
and the longitudinal distance travelled. Safety is assessed by the
proposed risk score shown in equation (\ref{eq:risk_score}) as follows:
\begin{equation}
\mathrm{Risk}=\frac{1}{T}\int_{0}^{T}U_{o}(t)\mathrm{dt},
\end{equation}
where $T$ is the planning horizon in seconds. Minimum jerk -- the
the time rate of change in acceleration -- reflects maximal smoothness
, and thus, the discomfort score can be defined using integrated absolute
jerk \cite{hogan2009sensitivity}:
\begin{equation}
\mathrm{Discomfort}=\frac{1}{T}\int_{0}^{T}\left|J(t)\right|\mathrm{dt},
\end{equation}
where, $J=\sqrt{J_{\mathrm{long}}^{2}+J_{\mathrm{lat}}^{2}}$ is the
total jerk experienced in the trajectory in both longitudinal and
lateral directions. The longitudinal distance is extracted from the
trajectory itself.
\begin{figure*}[t]
\begin{centering}
\includegraphics[scale=0.285]{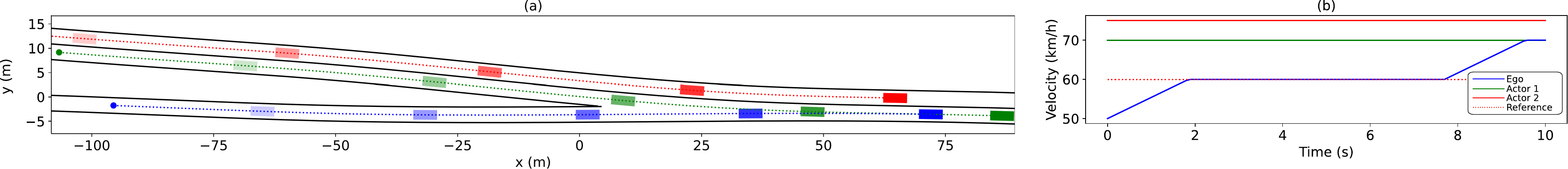}
\par\end{centering}
\caption{(a) The ego (blue rectangle) merging into the highway from the start (dot)
and the rectangles represent the progression of the vehicles over
time spaced by 2 seconds. (b) The velocity profiles of the ego and
the actors for the merging task \label{fig:Merging}}
\end{figure*}
\begin{figure*}[t]
\begin{centering}
\includegraphics[scale=0.285]{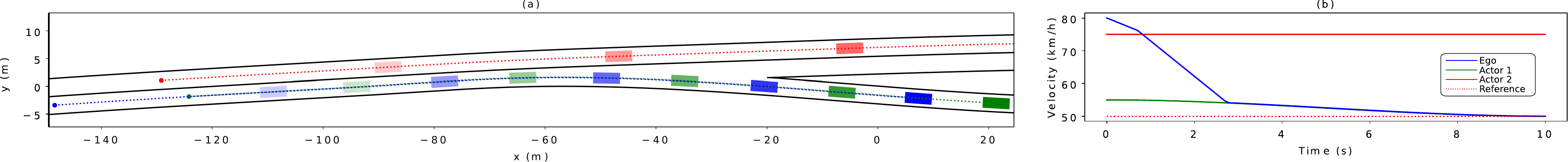}
\par\end{centering}
\caption{(a) The ego (in blue) taking the exit and the rectangles represent the
progression of the vehicles over time spaced by 2 seconds. (b) The
velocity profiles of the ego and the actors for the exit task \label{fig:Exit}}
\end{figure*}

\subsection{Experimental Results}

For the task of driving through medium and high density traffics,
our STG planner achieved 100\% success rate in planning feasible trajectories,
i.e., trajectories that do not cause a collision and do not violate
the kinematic constraints. However, MFP failed to find feasible trajectories
in 27\%, 29\%, and 33\% of the scenarios with low, medium, and high
traffic, respectively. The SUMO2015 model being the basis to generate
the trajectories, does not account for any failure. Table \ref{tab:Performance-comparison}
shows the performance comparisons between the three planners for the
driving through traffic task. The median score is presented for each
of the metrics as it defines the central tendency. Note that the results
for MFP shown in the table include only the scenarios where it succeeded. 

For all types of traffic, MFP generates the most comfortable trajectories
and the proposed planner stands second in the same aspect. However,
the primary objective of a planner is the ability to find a feasible
path, and our proposed STG planner demonstrates 100\% success rate
in planning in contrast to MFP. In low-density traffic scenarios,
MFP performs the best in terms of risk, while SL2015 performs the
best in terms of longitudinal distance travelled. However, STG achieves
a better risk score than SL2015 and has a better longitudinal distance travelled
than MFP for the same traffic. The scores for medium and high-density
the traffic shows that our planner achieves not only more longitudinal
distance but also safer trajectories. MFP generates trajectories
that are the shortest in terms of longitudinal distance, while SL2015
generates trajectories that are the least safe.
\begin{table}[t]
\caption{Median score comparisons (best score in bold) between the proposed
and the baselines (100 random scenarios for each type of traffic)
\label{tab:Performance-comparison}}

\centering{}%
\begin{tabular}{|>{\raggedright}p{0.65in}|c|>{\centering}p{0.55in}|c|>{\centering}p{0.55in}|}
\hline 
\textbf{\scriptsize{}Traffic Type} & \textbf{\scriptsize{}Planner} & \textbf{\scriptsize{}Discomfort {[}$\mathrm{m/s^{3}}${]}} & \textbf{\scriptsize{}Risk} & \textbf{\scriptsize{}Longitudinal Distance {[}$\mathrm{m}${]}}\tabularnewline
\hline 
\multirow{3}{0.65in}{\textbf{\scriptsize{}Low density traffic}} & \textbf{\scriptsize{}STG} & {\scriptsize{}0.52} & {\scriptsize{}5.76} & {\scriptsize{}108.24}\tabularnewline
\cline{2-5} \cline{3-5} \cline{4-5} \cline{5-5} 
 & \textbf{\scriptsize{}MFP{*}} & \textbf{\scriptsize{}0.2} & \textbf{\scriptsize{}5.19} & {\scriptsize{}104.22}\tabularnewline
\cline{2-5} \cline{3-5} \cline{4-5} \cline{5-5} 
 & \textbf{\scriptsize{}SL2015} & {\scriptsize{}2.41} & {\scriptsize{}5.81} & \textbf{\scriptsize{}109.58}\tabularnewline
\hline 
\multirow{3}{0.65in}{\textbf{\scriptsize{}Medium density traffic}} & \textbf{\scriptsize{}STG} & {\scriptsize{}1.04} & \textbf{\scriptsize{}6.30} & \textbf{\scriptsize{}108.55}\tabularnewline
\cline{2-5} \cline{3-5} \cline{4-5} \cline{5-5} 
 & \textbf{\scriptsize{}MFP{*}{*}} & \textbf{\scriptsize{}0.22} & {\scriptsize{}7.12} & {\scriptsize{}96.85}\tabularnewline
\cline{2-5} \cline{3-5} \cline{4-5} \cline{5-5} 
 & \textbf{\scriptsize{}SL2015} & {\scriptsize{}2.26} & {\scriptsize{}10.24} & {\scriptsize{}96.98}\tabularnewline
\hline 
\multirow{3}{0.65in}{\textbf{\scriptsize{}High density traffic}} & \textbf{\scriptsize{}STG} & {\scriptsize{}1.12} & \textbf{\scriptsize{}7.08} & \textbf{\scriptsize{}109.39}\tabularnewline
\cline{2-5} \cline{3-5} \cline{4-5} \cline{5-5} 
 & \textbf{\scriptsize{}MFP{*}{*}{*}} & \textbf{\scriptsize{}0.26} & {\scriptsize{}7.42} & {\scriptsize{}97.92}\tabularnewline
\cline{2-5} \cline{3-5} \cline{4-5} \cline{5-5} 
 & \textbf{\scriptsize{}SL2015} & {\scriptsize{}2.88} & {\scriptsize{}10.84} & {\scriptsize{}98.26}\tabularnewline
\hline 
\multicolumn{5}{l}{{\scriptsize{}{*}73 feasible trajectories, {*}{*}71, {*}{*}{*}67}}\tabularnewline
\end{tabular}
\end{table}
Figure \ref{fig:Merging} shows the performance of STG in the merging
task. The velocity profile shows that it accelerates to reach the
recommended speed ($\textrm{60 km/h}$) to merge. It continues with
the speed until it gets behind a lead actor (in green) and starts
following it effectively. In the task of taking an exit, STG generates
a safe trajectory, as well as shown in Figure \ref{fig:Exit}. The
ego starts to slow down but finds a lead actor (in green also taking
the exit) and then follows it quite efficiently to take the exit.
In both the above two tasks, STG demonstrates its efficacy in the
following three sub-tasks: speed-keeping, car-following and lane-keeping.
\begin{figure}[t]
\begin{centering}
\includegraphics[scale=0.16]{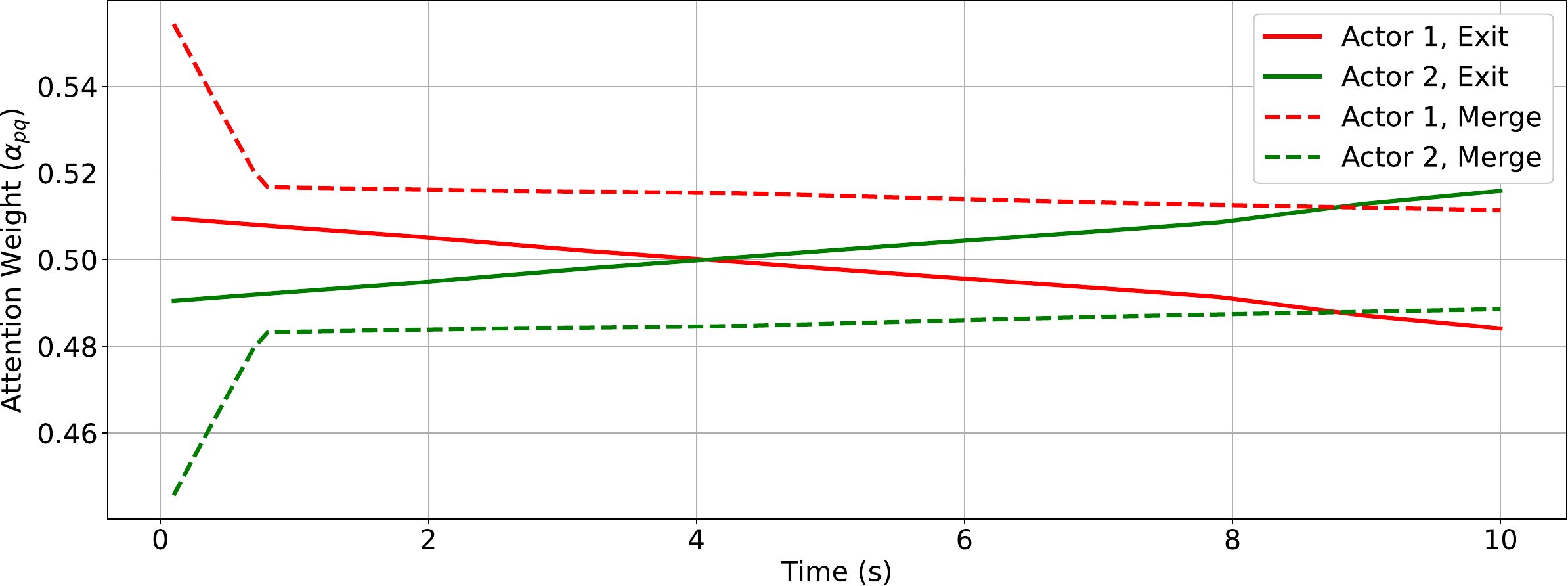}
\par\end{centering}
\caption{$\alpha_{pq}$ between ego $p$ and actor $q$\label{fig:-Attention_Weights}}
\end{figure}

\subsection{Interpretability}

The trained GAT network encapsulates the interactions between the
nodes with an attention mechanism as explained in section \ref{subsec:Graph-Neural-Network}.
Figure \ref{fig:-Attention_Weights} shows the attention in equation
(\ref{eq:attention_coeff}) evolving over time between the ego and
the actors in the tasks defined in sections \ref{subsec:Merging}
and \ref{subsec:Taking-an-Exit}. In the merging task, it can be seen
that since both the actors are closing in and remain close, their
attention values tend to converge towards each other. In the other
task of taking an exit, the attention of the actor taking the exit
ahead of the ego increases while the attention of the actor not taking
the exit decreases. While it is difficult to explain the magnitude
of these attention values, it is rather intuitive that the ego should
have increasing focus on the actor taking exit along with it rather than
the one going in another direction.

\section{Conclusion \label{sec:Conclusion} }

In this work, a novel online spatial-temporal graph trajectory planner
is introduced. Heterogeneous graphs are used to formulate the problem
by taking into account road boundaries and kinematic constraints. Then,
these graphs are processed in a sequential neural network architecture
to get the desired future trajectory \textcolor{black}{that can depict
common driving behaviors such as lane-keeping, lane-changing, car-following,
and speed-keeping. }For the network to learn, potential functions
addressing safety and maximum velocity keeping are presented as well.
\textcolor{black}{In addition, a simple behavioral layer is presented
to provide kinematic constraints for the planner. }The results show
that our proposed planner succeeded in generating feasible trajectories
in all the driving tasks. In contrast to the baselines, the metrics
point to better trade-off performance.

In the future, we want to test the efficacy of the proposed planner
in more diverse driving tasks using more complex behavioral layers.
We also plan to extend the application of our method to cooperative
multi-agent connected autonomous driving.

\appendix{}

\noindent 
\begin{algorithm}[H]
\caption{\textcolor{black}{Pseudo code to obtain kinematic constraints}\textcolor{blue}{\label{alg:Kinematic Constraints}}}

\begin{lyxcode}
\textbf{\scriptsize{}function~Kinematic~Constraints~}{\scriptsize\par}

\textbf{\scriptsize{}Inputs}{\scriptsize{}:~}{\scriptsize\par}
\begin{lyxcode}
{\scriptsize{}$s_{\mathrm{lead}},$~$s_{\mathrm{rear}},$~$v_{\mathrm{lead}},$~$v_{\mathrm{rear}},$~$a_{\mathrm{max,long}},$~$v_{\mathrm{max}},$~}{\scriptsize\par}

{\scriptsize{}$v_{\mathrm{min}},$~$a_{\mathrm{max,lat}},$~$v_{\mathrm{rec}},$~$s_{\mathrm{safe}}$,~DTT,~FSPS}{\scriptsize\par}
\end{lyxcode}
\textbf{\scriptsize{}Do}{\scriptsize{}:}{\scriptsize\par}
\begin{lyxcode}
{\scriptsize{}1~~if~not$\left(s_{\mathrm{lead}}<s_{\mathrm{safe}}\textrm{ or }s_{\mathrm{rear}}<s_{\mathrm{safe}}\right)$}{\scriptsize\par}

{\scriptsize{}2~~$\qquad\ddot{s}_{\mathrm{dec,max}}=a_{\mathrm{max,long}}$}{\scriptsize\par}

{\scriptsize{}3~~$\qquad\ddot{s}_{\mathrm{acc,max}}=a_{\mathrm{max,long}}$}{\scriptsize\par}

{\scriptsize{}4~~$\qquad\dot{s}_{\mathrm{max}}=v_{\mathrm{max}}$}{\scriptsize\par}

{\scriptsize{}5~~$\qquad\dot{s}_{\mathrm{min}}=v_{\mathrm{min}}$}{\scriptsize\par}

{\scriptsize{}6~~$\qquad\ddot{d}_{\mathrm{max}}=a_{\mathrm{max,lat}}$}{\scriptsize\par}

{\scriptsize{}\vspace{0.2in}
}{\scriptsize\par}

{\scriptsize{}7~~if~$s_{\mathrm{lead}}<s_{\mathrm{safe}}$}{\scriptsize\par}

{\scriptsize{}8~~$\qquad$$\ddot{s}_{\mathrm{dec,max}}=2a_{\mathrm{max,long}}$}{\scriptsize\par}

{\scriptsize{}9~~$\qquad$$\dot{s}_{\mathrm{max}}=v_{\mathrm{lead}}$}{\scriptsize\par}

{\scriptsize{}10~$\qquad$if~FSPS}{\scriptsize\par}

{\scriptsize{}11~$\qquad\qquad$$v_{\mathrm{rec}}=v_{\mathrm{lead}}$}{\scriptsize\par}

{\scriptsize{}\vspace{0.2in}
}{\scriptsize\par}

{\scriptsize{}12~if~$s_{\mathrm{rear}}<s_{\mathrm{safe}}$}{\scriptsize\par}

{\scriptsize{}13~$\qquad$$\ddot{s}_{\mathrm{acc,max}}=2a_{\mathrm{max,long}}$}{\scriptsize\par}

{\scriptsize{}14~$\qquad$$\dot{s}_{\mathrm{min}}=v_{\mathrm{rear}}$}{\scriptsize\par}

{\scriptsize{}15~$\qquad$if~FSPS~and~$v_{\mathrm{rec}}<\dot{s}_{\mathrm{min}}$}{\scriptsize\par}

{\scriptsize{}16~$\qquad\qquad$$v_{\mathrm{rec}}=\dot{s}_{\mathrm{min}}$}{\scriptsize\par}

{\scriptsize{}\vspace{0.2in}
}{\scriptsize\par}

{\scriptsize{}17~if~$s_{\mathrm{lead}}<s_{\mathrm{safe}}\textrm{ and }s_{\mathrm{rear}}<s_{\mathrm{safe}}$}{\scriptsize\par}

{\scriptsize{}18~$\qquad$$\ddot{d}_{\mathrm{max}}=2a_{\mathrm{max,lat}}$}{\scriptsize\par}

{\scriptsize{}19~$\qquad$if~$\dot{s}_{\mathrm{min}}>\dot{s}_{\mathrm{max}}$}{\scriptsize\par}

{\scriptsize{}20~$\qquad\qquad$$\dot{s}_{\mathrm{max}}=\dot{s}_{\mathrm{min}}$}{\scriptsize\par}
\end{lyxcode}
\textbf{\scriptsize{}Return}{\scriptsize{}:}{\scriptsize\par}

{\scriptsize{}~~~~if~DTT:~$\ddot{s}_{\mathrm{dec,max}},$~$\ddot{s}_{\mathrm{acc,max}},$~$\dot{s}_{\mathrm{max}},$~$\dot{s}_{\mathrm{min}},$~$\ddot{d}_{\mathrm{max}}$}{\scriptsize\par}

{\scriptsize{}~~~~if~FSPS:~$\ddot{s}_{\mathrm{dec,max}},$~$\ddot{s}_{\mathrm{acc,max}},$~$\ddot{d}_{\mathrm{max}},$~$v_{\mathrm{rec}}$}{\scriptsize\par}
\end{lyxcode}
\end{algorithm}
\textcolor{black}{Routine \ref{alg:Kinematic Constraints} shows the
steps to obtain the kinematic constraints. Lines 1-6 imply that in
the absence of a lead actor or a rear actor within the safety gap,
the kinematic constraints should remain the same, which were determined
prioritizing comfort. However, if there is a lead actor within the
safety gap (line 7), then the maximum deceleration rate $\ddot{s}_{\mathrm{dec,max}}$
is doubled (line 8) and the maximum velocity $\dot{s}_{\mathrm{max}}$
is constrained to lead actor's velocity $v_{\mathrm{lead}}$ (line
9). If the task is FSPS, the recommended speed $v_{\mathrm{rec}}$
is set to $v_{\mathrm{lead}}$ (line 11). Similarly, if there is a
rear actor within the safety gap (line 12), the maximum acceleration
rate $\ddot{s}_{\mathrm{acc,max}}$ is doubled (line 13) and the minimum
velocity $\dot{s}_{\mathrm{min}}$ is constrained to rear actor's
velocity $v_{\mathrm{rear}}$ (line 14). With FSPS, the recommended
velocity $v_{\mathrm{ref}}$ is set to $\dot{s}_{\mathrm{min}}$ if
$v_{\mathrm{ref}}<\dot{s}_{\mathrm{min}}$ (line 16). In the event
that there is both a lead and a rear actor within the safety gap (line
17), the lateral acceleration $\ddot{d}_{\mathrm{max}}$ is doubled
(line 18). Furthermore, the maximum velocity is adjusted in the event
that the rear actor is moving faster than the lead actor (lines 19-20).}

\textcolor{black}{It is important to note that doubling both the lateral
and longitudinal accelerations from the already given acceleration
constraints is a heuristic for safety -- the ego should be able to
speed up its way out of danger in the event a lead or a rear actor
breaches the safety gap without losing control. For example, with
a higher lateral acceleration in effect, the ego can swerve away faster
from the near-colliding vehicle, and this behavior is common in human
drivers \cite{li2019drivers}.}

\bibliographystyle{IEEEtran}
\bibliography{OnlineTrajectory_References}

\section*{Biography }

\vspace{-2pt}

\begin{IEEEbiography}[{\includegraphics[width=1in,height=1.25in]{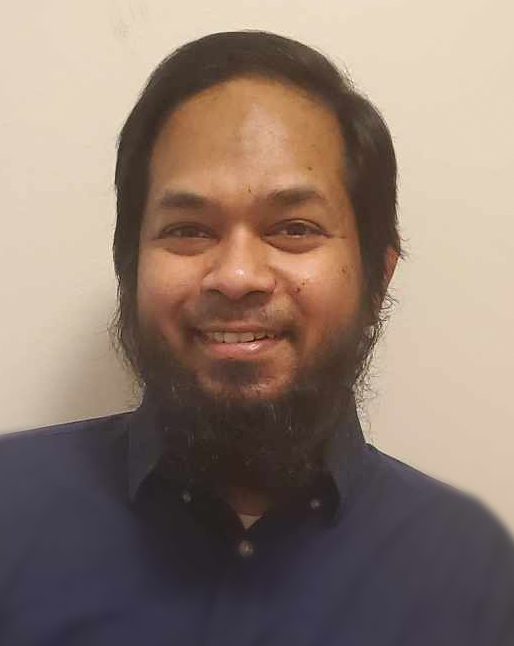}}]{Jilan Samiuddin}
  received his B.Sc. degree in Electrical and Electronic Engineering
from American International University - Bangladesh and M.Sc. degree
in Electrical Engineering from the University of Calgary in 2012 and
2016, respectively. He is currently a Ph.D. candidate in the Department
of Electrical and Computer Engineering at McGill University. His research
is primarily focused on machine learning algorithms with the motivation
of developing novel strategies for autonomous driving.
\end{IEEEbiography}

\vspace{-2pt}

\begin{IEEEbiography}[{\includegraphics[width=1in,height=1.25in]{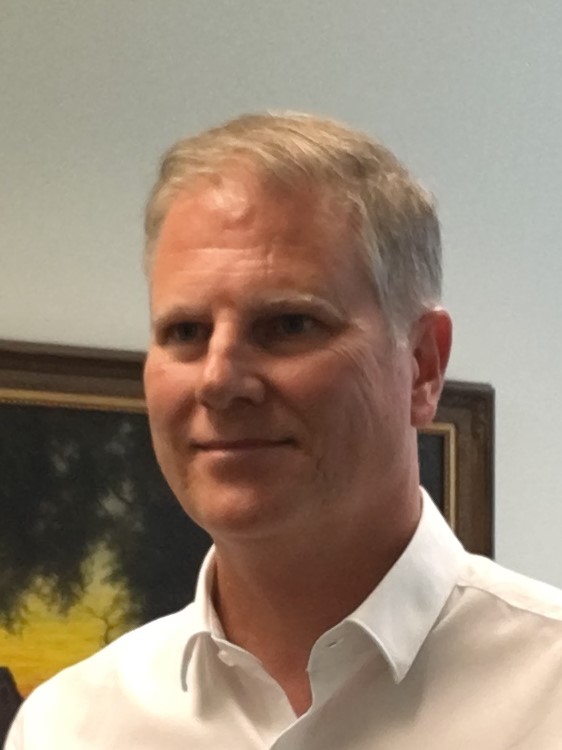}}]{Benoit Boulet}
 , P.Eng., Ph.D., is Professor of Electrical and Computer Engineering
at McGill University which he joined in 1998, and Director of the
McGill Engine. He is also McGill\textquoteright s Associate Vice-President
(Innovation and Partnerships). Prof. Boulet obtained an M.Eng. degree
from McGill in 1992 and a Ph.D. degree from the University of Toronto
in 1996, both in electrical engineering. He is a former Director and
current member of the Centre for Intelligent Machines where he heads
the Intelligent Automation Laboratory. His research interests include
the design and data-driven control of autonomous electric vehicles
and renewable energy systems and applications of machine learning.
\end{IEEEbiography}

\vspace{-2pt}

\begin{IEEEbiography}[{\includegraphics[width=1in,height=1.25in]{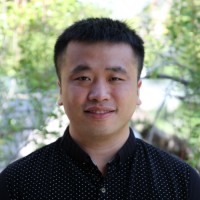}}]{Di Wu}
  is a senior staff research scientist, a team leader at Samsung
AI Center Montreal and an Adjunct Professor at McGill University since
2019. He did postdoctoral research at Montreal MILA and Stanford University.
He received his Ph.D. and M.Sc. from McGill University in 2018 and
Peking University in 2013, respectively. Di also holds Bachelor's
degrees in microelectronics and economics. His research interests
mainly lie in reinforcement learning, transfer learning, meta-Learning,
and multitask Learning. He is also interested in leveraging such algorithms
to improve real-world systems.
\end{IEEEbiography}

\end{document}